%% file: Formatting-Instructions-LaTeX-2026.tex
\documentclass[letterpaper]{article} % DO NOT CHANGE THIS
\usepackage{aaai2026}  % DO NOT CHANGE THIS
\usepackage{times}  % DO NOT CHANGE THIS
\usepackage{helvet}  % DO NOT CHANGE THIS
\usepackage{courier}  % DO NOT CHANGE THIS
\usepackage[hyphens]{url}  % DO NOT CHANGE THIS
\usepackage{graphicx} % DO NOT CHANGE THIS
\usepackage{natbib}  % DO NOT CHANGE THIS AND DO NOT ADD ANY OPTIONS TO IT
\usepackage{caption} % DO NOT CHANGE THIS AND DO NOT ADD ANY OPTIONS TO IT
\usepackage{algorithm}
\usepackage{algorithmic}

\usepackage{amsfonts}
\usepackage{amsmath}
\usepackage{booktabs}
\usepackage{multirow}
\usepackage{makecell}

\usepackage{newfloat}
\usepackage{listings}
\DeclareCaptionStyle{ruled}{labelfont=normalfont,labelsep=colon,strut=off} % DO NOT CHANGE THIS
\floatstyle{ruled}
\newfloat{listing}{tb}{lst}{}
\floatname{listing}{Listing}
\title{Class Incremental Medical Image Segmentation via \\ Prototype-Guided Calibration and Dual-Aligned Distillation}
\author{
    Shengqian Zhu\textsuperscript{\rm 1}, Chengrong Yu\textsuperscript{\rm 1}, Qiang Wang\textsuperscript{\rm 1}, Ying Song\textsuperscript{\rm 1}, Guangjun Li\textsuperscript{\rm 1}, \\
    Jiafei Wu\textsuperscript{\rm 2}, Xiaogang Xu\textsuperscript{\rm 3}\equalcontrib, Zhang Yi\textsuperscript{\rm 1}, Junjie Hu\textsuperscript{\rm 1}\equalcontrib \\
}
\affiliations{
    \textsuperscript{\rm 1}Sichuan University, Chengdu, Sichuan, China\\
    
    \textsuperscript{\rm 2}Zhejiang Lab, Hangzhou, Zhejiang, China\\
    
    \textsuperscript{\rm 3}The Chinese University of Hong Kong, Hong Kong, China\\
    zhushengqianscu@gmail.com
}

\usepackage{bibentry}
\begin{document}

\maketitle

\begin{abstract}
Class incremental medical image segmentation (CIMIS) aims to preserve knowledge of previously learned classes while learning new ones without relying on old-class labels. However, existing methods 1) either adopt one-size-fits-all strategies that treat all spatial regions and feature channels equally, which may hinder the preservation of accurate old knowledge, 2) or focus solely on aligning local prototypes with global ones for old classes while overlooking their local representations in new data, leading to knowledge degradation. To mitigate the above issues, we propose Prototype-Guided Calibration Distillation (PGCD) and Dual-Aligned Prototype Distillation (DAPD) for CIMIS in this paper. Specifically, PGCD exploits prototype-to-feature similarity to calibrate class-specific distillation intensity in different spatial regions, effectively reinforcing reliable old knowledge and suppressing misleading information from old classes. Complementarily, DAPD aligns the local prototypes of old classes extracted from the current model with both global prototypes and local prototypes, further enhancing segmentation performance on old categories. Comprehensive evaluations on two widely used multi-organ segmentation benchmarks demonstrate that our method outperforms state-of-the-art methods, highlighting its robustness and generalization capabilities.
\end{abstract}

\begin{links}
    \link{Code}{https://github.com/shengqianzhu/PCDD}
    \link{Extended version}{https://arxiv.org/abs/xxxx.xxxxx}
\end{links}

\section{Introduction}

Accurate medical image segmentation plays a vital role in a wide range of clinical applications, including diagnosis~\cite{zhang2017mdnet}, treatment planning~\cite{bauer2013survey, burgos2017iterative}, and particularly radiation therapy~\cite{qi2024gradient, zhu2025visual}. In radiotherapy, precise delineation of organs at risk and tumors is essential across multiple stages, such as dose calculation, beam optimization, and adaptive planning, to ensure both treatment efficacy and patient safety~\cite{wieser2017development}.

Traditionally, supervised deep learning models have achieved remarkable success in medical image segmentation by leveraging large-scale, fully annotated datasets~\cite{landman2015miccai, qi2023mdf, zhu2025exploiting}. However, in real-world clinical practice, new anatomical structures are typically introduced incrementally, while legacy data are inaccessible due to proprietary constraints, storage limitations, or practical limitations in reuse~\cite{michieli2019incremental}. In the absence of labels from previous classes, class incremental semantic segmentation (CISS) is required to learn new categories while retaining knowledge of previously learned ones to mitigate catastrophic forgetting~\cite{mccloskey1989catastrophic, wu2024continual}.

\begin{figure}[t]
\centering
\includegraphics[width=\columnwidth]{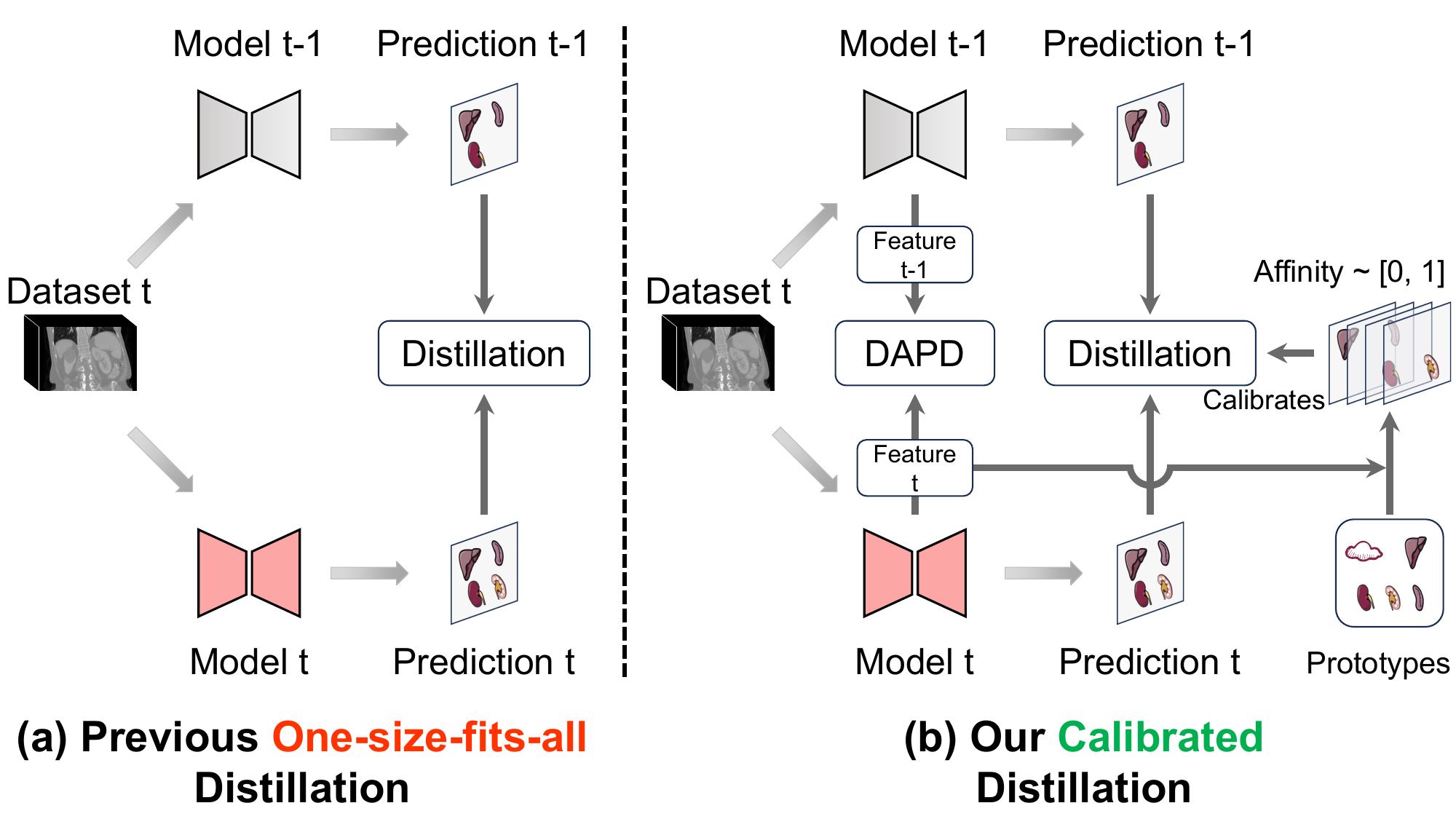} % Reduce the figure size so that it is slightly narrower than the column. Don't use precise values for figure width.This setup will avoid overfull boxes.
\caption{The conceptual comparison of (a) previous one-size-fits-all knowledge distillation and our (b) prototype-guided calibrated distillation. Unlike previous approaches that treated all regions and channels equally, our method calibrates the distillation process under the guidance of class prototypes, thereby better preserving useful old knowledge while suppressing misleading signals. The proposed DAPD further improves the segmentation performance of previously learned classes.}
\label{fig:first_graph}
\end{figure}

In order to mitigate forgetting of old classes, existing approaches typically resort to 1) knowledge distillation~\cite{michieli2019incremental, cermelli2020modeling, phan2022class}, while others leverage 2) replay-based techniques by retaining intermediate features~\cite{michieli2021continual, wu2023continual} (\textit{e.g.}, prototype, which is the representative feature vector of a class in the embedding space) from prior steps. Although these methods have achieved promising results, they still suffer from certain limitations in specific aspects. Specifically, existing distillation-based approaches often adopt a one-size-fits-all strategy~\cite{cermelli2020modeling}, applying identical distillation across the entire feature map without considering region-specific or channel-wise relevance. In old-class regions, each voxel may carry ambiguous or uncertain semantic information, particularly when different categories exhibit semantic proximity, making one-size-fits-all distillation prone to gradual forgetting of old knowledge. In new-class regions, one-size-fits-all distillation may introduce misleading supervision from incorrect old-class predictions, thereby impairing the model's ability to distinguish between old and new classes. On the other hand, the replay-based methods~\cite{michieli2021continual, wu2023continual} solely focus on aligning local prototypes with global ones for old classes, overlooking their local representations in new data (Left part of Fig.~\ref{fig:main_graph}). Moreover, these methods assume a fixed background prototype, which contradicts the principle that background semantics evolve across incremental learning steps due to changing class labels (\textit{i.e.}, background shift~\cite{cermelli2020modeling}).

In this paper, we propose Prototype-Guided Calibration Distillation (PGCD) and Dual-Aligned Prototype Distillation (DAPD) to address the aforementioned issues. The former calibrates distillation across different regions and channels under the guidance of prototypes, reinforcing reliable old knowledge while suppressing misleading signals to facilitate new class learning. The latter aligns the local prototypes of old classes extracted from the current model with both the global prototypes and the local prototypes, further enhancing the retention of old class knowledge.

% Specifically, PGCD introduces a prototype-guided similarity metric to recalibrate features across different spatial locations and channels in the distillation process.
% This design adaptively reinforces reliable knowledge in old-class regions while suppressing potentially misleading old knowledge in new-class areas, thereby ensuring the learning of new categories. 

Specifically, PGCD computes the affinity between each pixel feature and the class-specific prototypes to quantify the similarity between the feature and each semantic class. Since the prototype represents the class center in the feature space~\cite{xu2022general,snell2017prototypical}, the computed affinity can be used to recalibrate the distillation signals across different spatial locations and feature channels (Fig.~\ref{fig:first_graph}). This design adaptively reinforces reliable knowledge in old-class regions while suppressing potentially misleading old knowledge in new-class areas, thereby ensuring the learning of new categories. 

Additionally, our method recalculates the background prototype at each stage to address background shift issues within the prototype. Subsequently, DAPD performs unbiased dual alignment by aligning the current model’s local prototypes of old classes with both the previous step’s global prototypes and the old model’s local prototypes. The global prototype of an old class encodes static semantic knowledge, while local prototypes derived from the old model capture dynamic, batch-specific contextual cues in new data. Compared with previous methods that only align global prototypes of old classes, our proposed DAPD further aligns local prototypes that capture distributional shifts in new data, thereby better preserving old class knowledge.

The primary contributions of this work are threefold:

\begin{itemize}

    \item
    We propose a prototype-guided calibrated distillation strategy that adaptively recalibrates distillation across different regions and channels based on semantic similarity with class prototypes, thereby enhancing the retention of reliable knowledge while suppressing misleading legacy information.

    \item
    We design a dual-aligned prototype distillation mechanism to better preserve the knowledge of old classes.
   
    \item
    Extensive experiments on two public multi-organ segmentation benchmarks (BTCV and WORD) demonstrate that our method consistently outperforms existing state-of-the-art (SOTA) approaches under multiple class incremental settings (\textit{e.g.}, with +1.26 \% DSC improvement on the BTCV dataset using the 4-4 setting).
    
\end{itemize}

\section{Related Work}

\subsection{Class Incremental Learning}
Class Incremental Learning (CIL)~\cite{masana2022class, zhou2024class} aims to enable deep models to continually learn new categories without forgetting previously acquired knowledge. Catastrophic forgetting~\cite{mccloskey1989catastrophic, robins1995catastrophic, french1999catastrophic,tang2024incremental} is a critical challenge in CIL, where newly acquired knowledge disrupts or overwrites previously learned information due to the unavailability of the old class label. Existing CIL methods can be broadly categorized into three main strategies: regularization-based, replay-based, and architectural-based approaches. Regularization-based approaches mitigate forgetting by constraining parameter updates, typically via knowledge distillation losses~\cite{li2017learning, douillard2020podnet} or importance-based penalties such as Elastic Weight Consolidation (EWC)~\cite{kirkpatrick2017overcoming}. Replay-based methods alleviate forgetting by storing and rehearsing a subset of past samples~\cite{bang2021rainbow, aljundi2019gradient}, synthetic data~\cite{shin2017continual}, or intermediate representations~\cite{iscen2020memory} during training. Architecture-based approaches design class-specific sub-networks that are progressively extended as new categories emerge~\cite{yan2021dynamically, wang2022foster,xu2021ddcat,xu2022mtformer,zhou2022model}. Class incremental Semantic Segmentation (CISS) extends the aforementioned CIL methods to the pixel-level prediction task, where models are required to adapt to new categories while preserving segmentation performance on previously seen classes. In this work, we primarily focus on two representative approaches in CISS: knowledge distillation-based and prototype replay-based methods.

\begin{figure*}[ht]
\centering
\includegraphics[width=\textwidth]{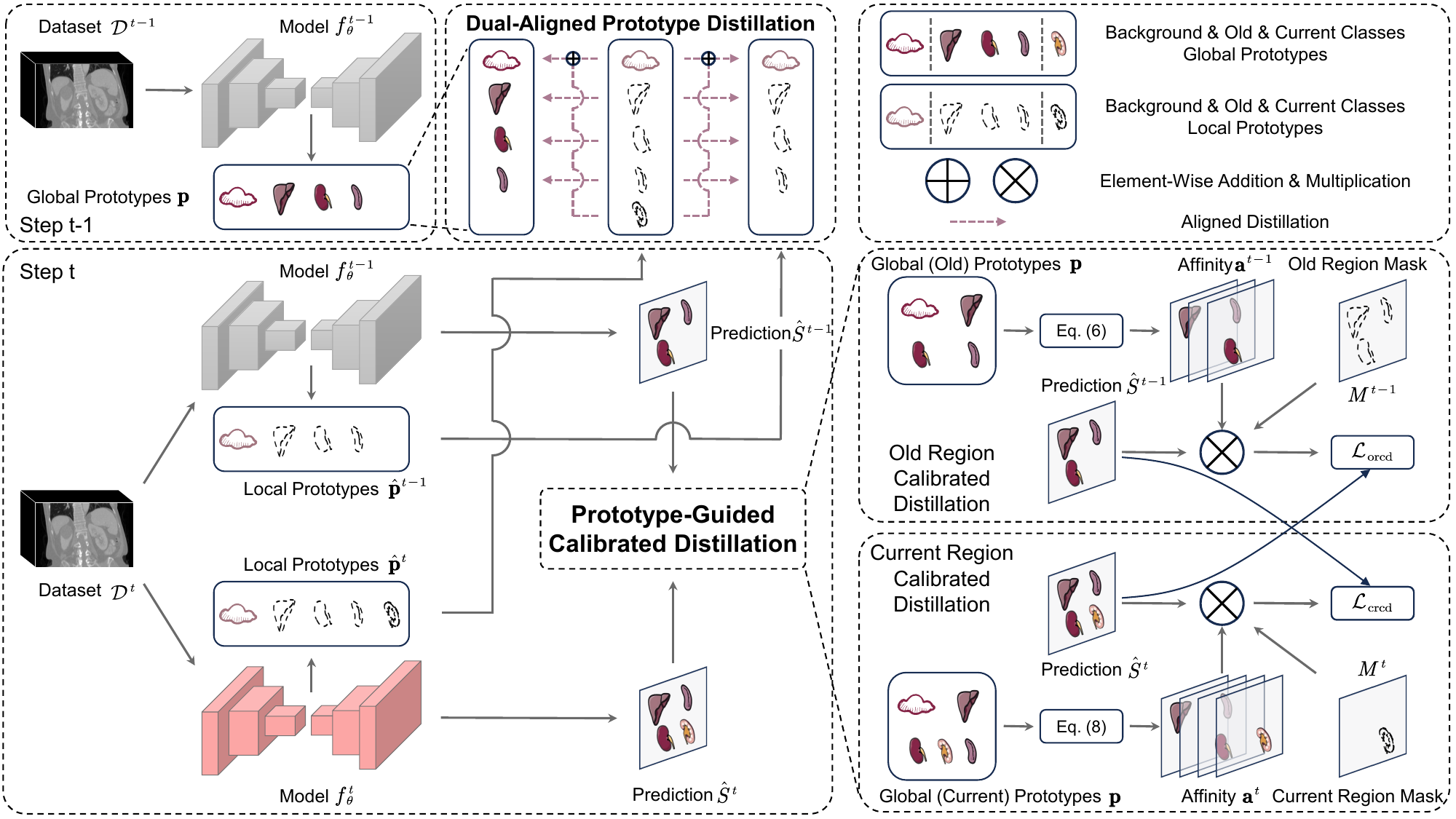} % Reduce the figure size so that it is slightly narrower than the column.
\caption{Overview of our proposed method. At step $t$, the model $f_{\theta}^{t}$ learns a new class (left kidney) while retaining knowledge of three old classes (liver, right kidney, and spleen). As shown in the top-left subfigure, global prototypes $\left \{ \mathbf{p}_c \mid c \in \mathcal{C}^{1:t-1} \cup 0 \right \} $ for the old classes are extracted and stored after step $t-1$. The bottom-left subfigure illustrates that, at the current step $t$, both the frozen model $f_{\theta}^{t-1}$ and the current model $f_{\theta}^{t}$ extract local prototypes $\left \{ \mathbf{\hat{p}}_c^{t-1} \mid c \in \mathcal{C}^{1:t-1} \cup 0 \right \} $ and $\left \{ \mathbf{\hat{p}}_c^{t} \mid c \in \mathcal{C}^{1:t} \cup 0 \right \} $, respectively. DAPD simultaneously aligns $\mathbf{\hat{p}}_c^{t}$ with both $\mathbf{\hat{p}}_c^{t-1}$ and $\mathbf{{p}}_c$ to enhance knowledge distillation for old classes. PGCD calibrates the distillation process across spatial regions and feature channels under the guidance of class prototypes.}
\label{fig:main_graph}
\end{figure*}

\subsubsection{Knowledge Distillation-Based CISS.}
ILT~\cite{michieli2019incremental}, one of the earliest works in CISS, aims to retain the old model’s knowledge by transferring its output layer predictions and intermediate representations to the new model via distillation. ILT performs distillation only on old-class outputs, neglecting that the background class can implicitly contain future-class semantics, which causes a background shift problem. MiB~\cite{cermelli2020modeling} introduces an unbiased output-level distillation strategy to address the background shift problem. Recent approaches~\cite{lin2022continual, zhang2024background, zhu2025prime} have introduced various improvements to intermediate feature distillation to better preserve knowledge of previously learned classes. However, these approaches typically adopt a one-size-fits-all distillation strategy, treating all feature channels and spatial locations equally, which leads to gradual forgetting of old class knowledge and hinders the learning of new classes. In contrast, our proposed PGCD adaptively calibrates distillation across different spatial regions and feature channels, thereby enhancing the retention of old knowledge while mitigating interference with new class learning.

\subsubsection {Prototype Replay-Based CISS.}

% replay -- SDR, CoNuSeg, InSeg
% michieli2021continual, wu2023continual, wang2024incremental
% Adaptive Prototype Replay for Class Incremental Semantic Segmentation

Prototype replay methods aim to regularize and enhance the feature representation learning in the latent space by replaying stored class prototypes. The class prototype (or centroid) serves as a compact and representative feature vector of a class in the latent feature space~\cite{snell2017prototypical}. For instance, SDR~\cite{michieli2021continual} leverages prototype matching to preserve the spatial structure of old classes. CoNuSeg~\cite{wu2023continual} leverages contrastive learning between class prototypes to enhance intra-class compactness and inter-class separability of features. InSeg~\cite{wang2024incremental} exploits prototypes to infer possible future categories and reinforce previously acquired feature representations. Adapter~\cite{zhu2025adaptive} introduces adaptive prototypes to robustly augment prototype replay in response to evolving class-level feature distributions during incremental learning. Nonetheless, these methods overlook the alignment of old classes in new data with both global and local prototypes simultaneously, and suffer from background shift issues in prototype representations. Unlike prior methods, we recompute the background prototype at each step, and DAPD performs an unbiased dual alignment to preserve the segmentation performance of old classes.

% applications\looseness=-1.

\section{Method}

\subsection{Problem Definition}

We focus on the task of CISS in the context of 3D medical image segmentation. Given a sequence of incremental learning steps $t = 1, \cdots, T$, the model is required to learn to segment a new set of semantic classes $\mathcal{C}^t$ at each step $t$, while retaining the ability to accurately segment all previously learned classes $\mathcal{C}^{1:t-1}$. At step $t$, the model receives a training dataset $\mathcal{D}^t = \left\{\left(\mathbf{X}_n, \mathbf{Y}_n \right) \right \}_{n=1}^{N_t}$, where $\mathbf{X}_n \in \mathbb{R}^{H \times W \times D}$ is a 3D medical image volume (\textit{e.g.}, CT or MRI), and $\mathbf{Y}_n \in \mathbb{L}^{H \times W \times D}$ is the corresponding voxel-wise segmentation label. The label space $\mathbb{L}$ includes only the current step's new classes $\mathcal{C}^t$, while both previously learned old classes $\mathcal{C}^{1:t-1}$ and future classes $\mathcal{C}^{t+1:T}$ are treated as background. Crucially, the datasets from previous steps $\mathcal{D}^{1:t-1}$ are assumed to be \textit{unavailable} due to privacy, storage, or legal constraints, which is a common scenario in clinical medical imaging applications. The objective is to train a segmentation model $f_{\theta}^{t}$ such that it can correctly segment all accumulated classes:

\begin{equation}
\begin{aligned}
    f_{\theta}^{t}\left(\mathbf{X} \right) \rightarrow \mathbf{\hat{Y}} \in \mathbb{R}^{ \left( \mathcal{C}^{1:t} + 1 \right) \times H \times W \times D},
    \label{eqn:model}
\end{aligned}
\end{equation} while avoiding catastrophic forgetting of old classes $\mathcal{C}^{1:t-1}$ and ensuring the model adapts effectively to new classes $\mathcal{C}^t$.

% To enhance readability, a comprehensive notation table is provided in the supplementary materials.

\subsection{Overview}

An overview of our proposed method is illustrated in Fig.~\ref{fig:main_graph}. At step $t$ of the incremental setting, the entire framework is composed of a frozen old model $f_{\theta}^{t-1}$ and a current model $f_{\theta}^{t}$ awaiting training. Our framework consists of two key components: PGCD and DAPD. PGCD comprises two components: Old Region Calibrated Distillation (ORCD) and Current Region Calibrated Distillation (CRCD), each focusing on calibrating the distillation of distinct spatial regions. DAPD adopts an unbiased dual-path prototype alignment strategy to enhance performance retention for previously learned classes. We elaborate on these individual components in the subsequent sections.

\subsection{Prototype Calculation}

To effectively guide the knowledge distillation process, we first compute class-wise prototypes that represent the semantic centers of each category in the feature space~\cite{snell2017prototypical}. Specifically, we maintain two types of prototypes during training: local prototypes (computed from the current batch) and global prototypes (updated across batches via a cumulative moving average). Given a mini-batch of features $\mathbf{F} \in \mathbb{R}^{B \times K \times H \times W \times D}$ extracted from an intermediate layer of the backbone network $f_{\theta}$, where $B$ and $K$ denote the batch size and channel dimension. Subsequently, we derive the present class set $\mathcal{C}$ from the corresponding label map $\mathbf{Y} \in \mathbb{R}^{B \times 1 \times H \times W \times D}$. 

For each class $c \in \mathcal{C}$, we extract the corresponding class-specific feature vectors: $\mathbf{F}_{c}=\left\{\mathbf{f}_{i} \in \mathbf{F} \mid \mathbf{Y}(i)=c\right\}$. The local prototype $\mathbf{\hat{p}}_c \in \mathbb{R}^K$ is then computed as the mean of those feature vectors:
\begin{equation}
\begin{aligned}
    \mathbf{\hat{p}}_c=\frac{1}{\left|\mathbf{F}_{c}\right|} \sum_{\mathbf{f}_{i} \in \mathbf{F}_{c}} \mathbf{f}_{i}.
    \label{eqn:local_pro}
\end{aligned}
\end{equation} 
To ensure robustness across the training set, the global prototype $\mathbf{p}_c \in \mathbb{R}^K$ for each class $c$ is progressively updated by a moving average strategy. Let $N_c^{\text{pre}}$ be the number of previously observed pixels for class $c$, and $\mathbf{p}_c^{\text{pre}}$ be the cumulative global prototypes so far. After observing a new mini-batch $B$ with feature vectors $\mathbf{F}_c^{B}$, the global prototype $\mathbf{p}_c$ and the number of currently observed pixels $N_c$ are updated as:

\begin{equation}
\begin{aligned}
    \mathbf{p}_c = \frac{N_c^{\text{pre}} + \left|\mathbf{F}_{c}^{B} \right| \cdot \sum_{\mathbf{f}_i \in \mathbf{F}_c^{B}} \mathbf{f}_i }{N_c^{\text{pre}} + \left|\mathbf{F}_{c}^{B} \right|}, \quad N_c = N_c^{\text{pre}} + \left|\mathbf{F}_{c}^{B} \right|.
    \label{eqn:global_pro}
\end{aligned}
\end{equation} 
This update rule ensures that the global prototype reflects a running average of all observed features per class, capturing the accurate semantics of each category during the long training process.

It should be noted that the prototype of the background class may incorporate latent features of future classes $\mathcal{C}^{t+1:T}$, as these are annotated as background in the ground truth. Unlike prior approaches that rely on a static background prototype, we recompute a new background prototype at each step to avoid impairing the segmentation performance of future classes. More details are elaborated in the subsequent DAPD section.

\subsection{Prototype-Guided Calibration Distillation}

Based on the prototypes, we propose PGCD to enhance reliable old knowledge while suppressing misleading signals, thereby ensuring the learning of new categories. Unlike conventional one-size-fits-all distillation strategies that treat all spatial regions and feature channels equally, PGCD adaptively calibrates knowledge transfer across different regions and channels using class-specific prototype guidance. 

First, we derive the pseudo labels based on the original design in PLOP~\cite{douillard2021plop}, as detailed below:

\begin{equation}
\begin{aligned}
    \tilde{\mathbf{Y}}_i =
    \begin{cases}
    \mathbf{Y}_i & \text{if } \mathbf{Y}_i \neq 0 \\
    \underset{c \in \mathcal{C}^{1:t-1}}{\arg \max }  \hat{\mathbf{S}}_{i,c}^{t-1} & \text{if } \underset{c}{\arg \max} \hat{\mathbf{S}}_{i,c}^{t-1} \neq 0, \text{ and } u_i < \tau \\
    0 & \text{otherwise}
    \end{cases},
    \label{eqn:pseudo_label}
\end{aligned}
\end{equation} where $\hat{\mathbf{S}}_{i,c}^{t-1}$ denote the softmax output of the old model $f_{\theta}^{t-1}$ at voxel $i$ for class $c \in \mathcal{C}^{1:t-1}$. Here, $u_i = -\sum_{c} \hat{\mathbf{S}}_{i,c}^{t-1} \log \hat{\mathbf{S}}_{i,c}^{t-1}$ represents the entropy of the softmax output at voxel $i$, serving as an uncertainty measure. The threshold $\tau$ is a pre-defined scalar that controls the acceptance of confident predictions.

After generating the pseudo labels $\tilde{\mathbf{Y}}_i$ (Eq.~\ref{eqn:pseudo_label}), we construct two spatial binary masks: the old region mask $\mathbf{M}_i^{t-1}$ and the new region mask $\mathbf{M}_i^{t}$, to facilitate region-specific knowledge distillation. These masks are computed as:

\begin{equation}
\mathbf{M}_i^{t-1} = \mathbb{I}\left[\tilde{Y}_i \in \mathcal{C}^{1:t-1}\right], \quad
\mathbf{M}_i^{t} = \mathbb{I}\left[\tilde{Y}_i \in \mathcal{C}^{t} \cup 0\right],
\label{eqn:region_masks}
\end{equation} 
where $\mathbb{I}[\cdot]$ is the indicator function. The old region mask identifies voxels confidently predicted as old classes $\mathcal{C}^{1:t-1}$, while the new region mask covers voxels that belong to either the new classes $\mathcal{C}^t$  or the background.

% At the current step $t$, the old model $f_{\theta}^{t-1}$ and the current model $f_{\theta}^{t}$ generate local prototypes for the old classes $\mathcal{C}^{1:t-1}$ and all classes $\mathcal{C}^{1:t}$ encountered so far, respectively.

\subsubsection{Old Region Calibrated Distillation.}
To adaptively enhance the reliable knowledge of old regions, we leverage the global prototypes $\left \{ \mathbf{p}_c \mid  \mathbf{p}_c \in \mathbb{R}^K, c \in \mathcal{C}^{1:t-1}  \right \} $ of old classes preserved from previous steps (1 to $t-1$) as guidance. We choose to use prototypes as guidance primarily for two reasons: 1) each prototype represents the center of a category and is thus representative and robust to outliers~\cite{cong2024cs}; 2) it is computed by averaging all feature vectors of a given class, thereby treating each category equally regardless of voxel frequency, which effectively mitigates the class imbalance issue in medical image segmentation.

Let $\mathbf{F}^{t-1} \in \mathbb{R}^{V \times K}$ be the feature map extracted from the old model $f_{\theta}^{t-1}$, where $V$ denotes the number of spatial locations. For each voxel $i$, we calculate its prototype-guided affinity vector $\mathbf{a}_i^{t-1} \in \mathbb{R}^{|\mathcal{C}^{1:t-1}|}$ by computing the cosine similarity between the feature at location $i$ and all old class global prototypes:

\begin{equation}
\begin{aligned}
\mathbf{a}_{i,c}^{t-1} = \frac{  \mathbf{F}_i^{t-1} \cdot \mathbf{p}_c  }{ \left\| \mathbf{F}_i^{t-1} \right\|_2 \left\| \mathbf{p}_c \right\|_2 }, \quad \forall c \in \mathcal{C}^{1:t-1},
\label{eq:affinity}
\end{aligned}
\end{equation} where $\left\| \cdot \right\|_2$ denote L2 norm (also known as the Euclidean norm). These affinities are then normalized (\textit{i.e.}, via softmax) to serve as voxel-wise class calibration weights in the distillation process.

Following the affinity calibration, we use these weights to calibrate the knowledge distillation from the old model’s prediction $\hat{\mathbf{S}}^{t-1} \in \mathbb{R}^{V \times |\mathcal{C}^{1:t-1}|}$ to the current model’s prediction $\hat{\mathbf{S}}^{t} \in \mathbb{R}^{V \times |\mathcal{C}^{1:t}|}$. Since $\hat{\mathbf{S}}^{(t)}$ contains logits for new classes $\mathcal{C}^{t}$ as well, we adopt the unbiased knowledge distillation strategy proposed in~\cite{cermelli2020modeling} to ensure the distillation is not biased towards the newly added background or classes. The ORCD loss is defined as:

\begin{equation}
\begin{aligned}
\mathcal{L}_{\text{orcd}} = \frac{1}{V^{t-1}} \sum_{i: M_i^{t-1}=1} \sum_{c \in \mathcal{C}^{1:t-1}} \mathbf{a}_{i,c}^{t-1} \cdot \text{KL} \left( \hat{\mathbf{S}}_{i,c}^{t} \parallel \hat{\mathbf{S}}_{i,c}^{t-1} \right),
\label{eq:ukd}
\end{aligned}
\end{equation} where $\text{KL} \left(\cdot \parallel \cdot \right)$ is the Kullback-Leibler divergence, and $V^{t-1}$ is the total number of voxels in the old region mask. 

\textbf{Rationale of ORCD.} In old class regions, each voxel may carry ambiguous or uncertain semantics due to inter-class similarity, rendering one-size-fits-all distillation prone to gradually forgetting previously learned knowledge. Voxels exhibiting higher similarity (\textit{i.e.}, higher affinity) to a class prototype are expected to receive stronger supervision signals in the corresponding class channel during distillation. The proposed ORCD reinforces reliable old knowledge and mitigates its potential degradation throughout the incremental learning steps.

\subsubsection{Current Region Calibrated Distillation.}

To mitigate the adverse impact of misleading knowledge from old classes on learning new categories, we propose CRCD, a mechanism that adaptively suppresses old knowledge within the current region $M_i^{t}$. We leverage global prototypes $\left \{ \mathbf{p}_c \mid \mathbf{p}_c \in \mathbb{R}^K, c \in \mathcal{C}^{1:t} \right \}$ of all seen classes to guide distillation. For each voxel $i$, we compute its prototype-guided affinity vector $\mathbf{a}_i^t$ as:

\begin{equation}
\mathbf{a}_{i,c}^{t} = \frac{ \mathbf{F}_i^t \cdot \mathbf{p}_c  }{ \left\| \mathbf{F}_i^t \right\|_2 \left\| \mathbf{p}_c \right\|_2 }, \quad \forall c \in \mathcal{C}^{1:t},
\label{eq:crcd_affinity}
\end{equation} where $\mathbf{F}^t \in \mathbb{R}^{V \times K}$ denotes the feature map of the current model $f_{\theta}^{t}$. The CRCD loss is:

\begin{equation}
\mathcal{L}_{\text{crcd}} = \frac{1}{V^{t}} \sum_{i: M_i^{t}=1} \sum_{c \in \mathcal{C}^{1:t}} \mathbf{a}_{i,c}^{t} \cdot \text{KL} \left( \hat{\mathbf{S}}_{i,c}^{t} \parallel \hat{\mathbf{S}}_{i,c}^{t-1} \right),
\label{eq:crcd}
\end{equation} where $V^t$ is the number of voxels in the current region. 

\textbf{Rationale of CRCD.} In the current region, some predictions may be mistakenly classified as old classes, and traditional distillation may lead the current model to imitate such incorrect predictions from the old model. The proposed CRCD reduces the distillation weight of incorrectly predicted old classes under the guidance of prototypes, thereby suppressing unreliable old predictions and preserving the model's ability to learn new categories.

\subsection{Dual-Aligned Prototype Distillation}

Complementary to the logit-based distillation discussed above, we design a DAPD approach to enhance knowledge transfer for old classes. Unlike conventional pixel-wise distillation, DAPD operates at the prototype level and enforces consistency between class-wise representations across the old and current models. Specifically, we align the local prototypes $\left \{ \mathbf{\hat{p}}_c^{t} \mid c \in \mathcal{C}^{1:t-1} \cup 0 \right \} $ of old classes (including background) extracted from the current model with both the local $\left \{ \mathbf{\hat{p}}_c^{t-1} \right \} $ and global $\left \{ \mathbf{p}_c \right \} $ prototypes of the corresponding classes from the old model. 

As previously discussed, to avoid the potential negative impact of a static background prototype on the segmentation performance of future classes, we recompute a new background prototype at each incremental step. This implies that, at the current step $t$, the background prototype ($\left \{ \mathbf{\hat{p}}_0^{t-1}, \mathbf{p}_0 \right \}$) of the old model encodes latent features of the current class $\mathcal{C}^{t}$. Before defining DAPD loss, we apply an unbiased transformation to the local prototypes derived from the current model as follows:

\begin{equation}
\begin{aligned}
    \mathbf{\hat{q}}_c = \left\{
   \begin{aligned}
    & {\textstyle \sum_{n \in \{0\} \cup \mathcal{C}^{t}}} \mathbf{\hat{p}}_c^{t}   & & \textrm{if } c = 0 \\
    & \mathbf{\hat{p}}_c^{t}                          & & \textrm{if } c \in \mathcal{C}^{1:t-1}  \\
   \end{aligned}
    \right..
    \label{eqn:main_merge}
\end{aligned}
\end{equation} Thereafter, the complete DAPD loss is defined as:
\begin{equation}
\begin{aligned}
\mathcal{L}_{\text{dapd}} = \boldsymbol{\lambda}_{\text{ll}}\mathcal{L}_{\text{pd}}\left ( \mathbf{\hat{q}}_c, \mathbf{\hat{p}}_c^{t-1} \right ) + \boldsymbol{\lambda}_{\text{lg}}\mathcal{L}_{\text{pd}}\left ( \mathbf{\hat{q}}_c, \mathbf{p}_c \right ),
\label{eq:dapd}
\end{aligned}
\end{equation} 
where $\boldsymbol{\lambda}_{\text{ll}}$ and $\boldsymbol{\lambda}_{\text{lg}}$ are the weights used to balance the contributions of each loss item. The loss term $\mathcal{L}_{\text{pd}} \left ( \mathbf{q}_c, \mathbf{p}_c \right ) $ represents the prototype distillation loss, which is defined as:
\begin{equation}
\begin{aligned}
    \mathcal{L}_{\text{pd}} \left ( \mathbf{\hat{q}}_c, \mathbf{p}_c \right )  = \frac{1}{|\mathcal{C}^{1:t-1}| + 1} \sum_{c \in \mathcal{C}^{1:t-1}\cup 0}  \left\| \mathbf{\hat{q}}_c - \mathbf{p}_c \right\|_2^2.
\label{eq:dapd}
\end{aligned}
\end{equation} By employing unbiased dual alignment, DAPD effectively retains old class knowledge.

\subsection{Overall Loss Function}

Based on the aforementioned loss components, the overall objective function of our method is defined as follows:

\begin{equation}
\begin{aligned}
    \mathcal{L}_{\text{total}} = \mathcal{L}_{\text{ce}} + \boldsymbol{\lambda}_{\text{orcd}} \mathcal{L}_{\text{orcd}} + \boldsymbol{\lambda}_{\text{crcd}} \mathcal{L}_{\text{crcd}} + \mathcal{L}_{\text{dapd}},
    \label{eqn:overall_loss}
\end{aligned}
\end{equation} where $\boldsymbol{\lambda}_{\text{orcd}}$ and $\boldsymbol{\lambda}_{\text{crcd}}$ are the weights used to adjust the impact of the corresponding loss on the total objective. The loss function $\mathcal{L}_{\text{ce}}$, implemented as the unbiased cross-entropy proposed in MiB~\cite{cermelli2020modeling}, supervises the learning of the current class.

\begin{table*}[ht]
\centering
%\resizebox{.95\columnwidth}{!}{
\begin{tabular}{c | ccc | ccc | ccc | ccc}
\toprule

     & \multicolumn{3}{c |}{4-4 (2 steps)} & \multicolumn{3}{c |}{4-2 (3 steps)} & \multicolumn{3}{c |}{4-1 (5 steps)} & \multicolumn{3}{c}{7-1 (2 steps)} \\ \cline{2-13}
Method & Old   & New   & All   & Old   & New   & All   & Old   & New   & All   & Old   & New   & All \\ \hline

Offline & 92.58  & 88.93  & 90.75  & 92.58  & 88.93  & 90.75  & 92.58  & 88.93  & 90.75  & 90.44  & 92.93  & 90.75  \\ \hline
LwF   & 85.77  & \underline{88.44}  & 87.11  & 1.50  & 73.30  & 37.40  & 0.00  & 47.06  & 23.53  & 65.67  & 92.93  & 69.08  \\
ILT   & 25.92  & 36.61  & 31.26  & 10.15  & 44.65  & 27.40  & 0.00  & 38.48  & 19.24  & 30.18  & 74.25  & 35.69  \\
MiB   & 91.27  & 86.96  & 89.12  & \textbf{92.51}  & 87.36  & \underline{89.94}  & \textbf{92.34}  & 88.24  & \underline{90.29}  & \underline{90.10}  & \underline{92.95}  & \underline{90.45}  \\
PLOP  & 57.37  & 37.12  & 47.24  & 86.72  & 57.40  & 72.06  & 67.26  & 25.69  & 46.47  & 78.77  & 77.30  & 78.58  \\
SDR   & \underline{92.07}  & 86.60  & \underline{89.34}  & 91.76  & \underline{87.47}  & 89.61  & 91.81  & \underline{88.31}  & 90.06  & 76.21  & 91.78  & 78.15  \\
REMINDER & 24.73  & 32.02  & 28.37  & 83.10  & 41.19  & 62.14  & 77.25  & 28.67  & 52.96  & 75.05  & 70.07  & 74.43  \\
CoNuSeg & 85.92  & 45.63  & 65.77  & 83.95  & 65.29  & 74.62  & 3.45  & 43.60  & 23.53  & 72.93  & 91.30  & 75.23  \\
InSeg & 55.83  & 37.15  & 46.49  & 88.75  & 61.66  & 75.21  & 71.49  & 43.71  & 57.60  & 73.66  & 74.30  & 73.74  \\ \hline
Ours  & \textbf{92.51} & \textbf{88.69} & \textbf{90.60} & \underline{92.45}  & \textbf{88.48 } & \textbf{90.46} & \underline{92.27}  & \textbf{88.83} & \textbf{90.55} & \textbf{90.15} & \textbf{93.37} & \textbf{90.55} \\

\bottomrule
\end{tabular}
\caption{Comparison with SOTA methods for different CISS tasks on the BTCV dataset using average DSC (\%). The best and the second results are marked in bold and underlined, respectively.}
\label{tab:main_result_BTCV}
\end{table*}

\begin{table*}[ht]
\centering
%\resizebox{.95\columnwidth}{!}{
\begin{tabular}{c | ccc | ccc | ccc | ccc}
\toprule

     & \multicolumn{3}{c |}{4-4 (2 steps)} & \multicolumn{3}{c |}{4-2 (3 steps)} & \multicolumn{3}{c |}{2-2 (4 steps)} & \multicolumn{3}{c}{7-1 (2 steps)} \\ \cline{2-13}
Method & Old   & New   & All   & Old   & New   & All   & Old   & New   & All   & Old   & New   & All \\ \hline

Offline & 86.18  & 84.77  & 85.47  & 86.18  & 84.77  & 85.47  & 94.43  & 82.49  & 85.47  & 84.92  & 89.34  & 85.47  \\ \hline
LwF   & 76.38  & \textbf{82.89} & 79.63  & 0.39  & 68.63  & 34.51  & 0.00  & 43.01  & 32.25  & 49.11  & \textbf{82.99} & 53.35  \\
ILT   & 61.80  & 36.93  & 49.36  & 9.40  & 43.03  & 26.22  & 0.04  & 15.54  & 11.66  & 25.57  & 13.30  & 24.03  \\
MiB   & \underline{83.88}  & 81.42  & \underline{82.65}  & 58.62  & 45.18  & 51.90  & \underline{91.79}  & \underline{64.26}  & \underline{71.14}  & \underline{71.05}  & 81.22  & \underline{72.32}  \\
PLOP  & 79.05  & 48.36  & 63.71  & \underline{82.75}  & 33.77  & 58.26  & 73.23  & 29.88  & 40.72  & 47.69  & 12.36  & 43.27  \\
SDR   & 81.28  & 81.58  & 81.43  & 80.22  & \underline{78.02}  & \underline{79.12}  & \textbf{92.29} & 57.73  & 66.37  & 70.59  & 81.26  & 71.92  \\
REMINDER & 68.22  & 45.04  & 56.63  & 78.23  & 41.67  & 59.95  & 27.91  & 23.75  & 24.79  & 35.15  & 11.29  & 32.16  \\
CoNuSeg & 78.18  & 63.43  & 70.80  & 0.08  & 20.94  & 10.51  & 43.16  & 47.12  & 46.13  & 57.57  & 75.77  & 59.85  \\
InSeg & 78.28  & 50.41  & 64.34  & 65.45  & 42.98  & 54.22  & 80.98  & 40.59  & 50.69  & 17.12  & 12.59  & 16.56  \\ \hline
Ours  & \textbf{84.09} & \underline{82.75}  & \textbf{83.42} & \textbf{83.95} & \textbf{83.04} & \textbf{83.50} & 90.17  & \textbf{65.97} & \textbf{72.02} & \textbf{71.51} & \underline{82.59}  & \textbf{72.89} \\

\bottomrule
\end{tabular}
\caption{Comparison with SOTA methods for different class incremental tasks on the WORD dataset using average DSC (\%). The best and the second results are marked in bold and underlined, respectively.}
\label{tab:main_result_WORD}
\end{table*}

\begin{figure*}[ht]
\centering
\includegraphics[width=\textwidth]{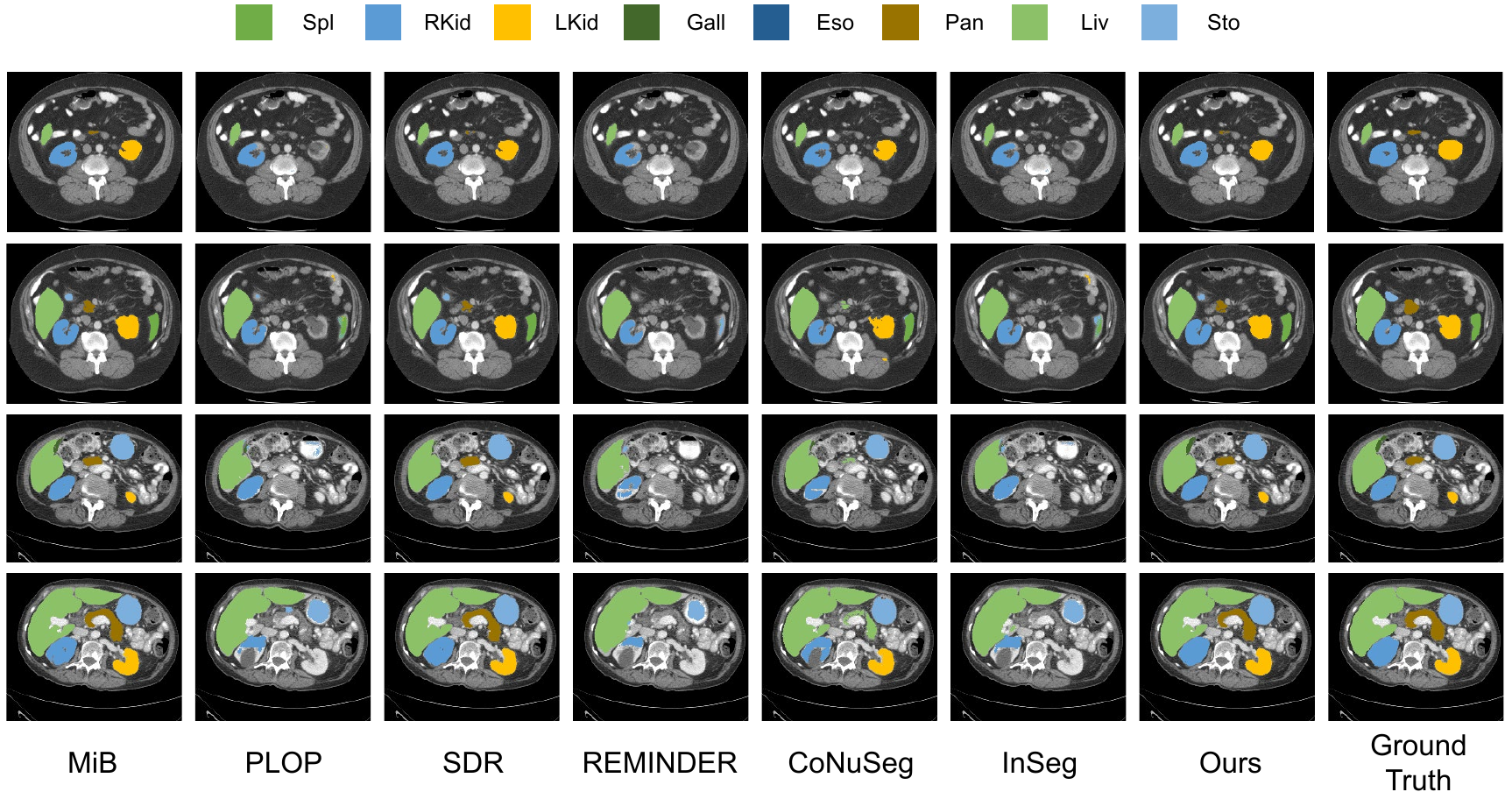} % Reduce the figure size so that it is slightly narrower than the column.
\caption{Qualitative comparison of segmentation results between the proposed method and SOTA approaches on BTCV 4-4.}
\label{fig:visualization}
\end{figure*}

\section{Experiments}
\subsection{Experimental Setup}

% originates from the \textit{MICCAI Multi-Atlas Labeling Beyond Cranial Vault Challenge} and
% \subsubsection{Dataset and Preprocessing.}
\subsubsection{Dataset.}
We evaluate our method on two public multi-organ segmentation datasets: BTCV~\cite{landman2015miccai} and WORD~\cite{luo2022word}. The BTCV dataset contains 30 abdominal CT scans with expert annotations for 13 abdominal organs. The WORD dataset is a large-scale CT dataset that includes 120 cases with annotations for 16 abdominal organs. To simplify the incremental learning setup across datasets, we focus on the 8 overlapping organs between these two datasets: spleen, right kidney, left kidney, gallbladder, esophagus, pancreas, liver, and stomach (which are also taken as the class order).

% The details of dataset preprocessing and implementation are provided in the supplementary materials.

\subsubsection{Protocols.}
We design several CISS protocols based on a unified notation, $N_1$-$N_2$ $(T , \text{steps})$, where $N_1$ and $N_2$ represent the number of classes introduced in the initial and each incremental step, respectively. The total number of steps $T$ is given by $1 + (|\mathcal{C}| - N_1) / N_2$, where $|\mathcal{C}|$ denotes the total number of classes. In our experiments, we consider the eight abdominal organs mentioned earlier, with their class order fixed consistently throughout all steps. In this way, we construct the following CISS protocols~\cite{wang2024incremental}: 4-4 (2 steps), 4-2 (3 steps), 2-2 (4 steps), 4-1 (5 steps), and 7-1 (2 steps), to assess the performance under different incremental tasks comprehensively.

\subsubsection{Evaluation Metrics.}
To evaluate our method in CISS, we use the Dice Similarity Coefficient (DSC) to measure the overlap between predictions and ground truth. At the final incremental step, we report: 1) Old: average DSC over initial classes ($\mathcal{C}^1$), assessing knowledge retention; 2) New: average DSC over newly added classes ($\mathcal{C}^{2:T}$), reflecting new knowledge acquisition; and 3) All: average DSC over all seen classes ($\mathcal{C}^{1:T}$), indicating overall segmentation performance and balance between retention and learning.

\subsubsection{Baselines.}
We compare our method against eight SOTA CISS approaches. These include 1) the regularization-based method LwF~\cite{li2017learning}, 2) distillation-based strategies:  ILT~\cite{michieli2019incremental}, MiB~\cite{cermelli2020modeling}, PLOP~\cite{douillard2021plop}, 3) prototype replay-based approaches: SDR~\cite{michieli2021continual}, REMINDER~\cite{phan2022class}, CoNuSeg~\cite{wu2023continual} and InSeg~\cite{wang2024incremental}. The \textit{offline} setting, where the model is trained on all classes simultaneously, serves as an upper bound for performance.

\begin{table}[ht]
\centering
%\resizebox{.95\columnwidth}{!}{
\begin{tabular}{cccc | ccc}
    \toprule
    $\boldsymbol{\lambda}_{\text{ll}}$ & $\boldsymbol{\lambda}_{\text{lg}}$ & $\boldsymbol{\lambda}_{\text{orcd}}$  & $\boldsymbol{\lambda}_{\text{crcd}}$ & Old & New & All \\
    \midrule
          &       &       &       & 91.27  & 86.96  & 89.12  \\
    \checkmark     &       &       &       & 92.11  & 87.43  & 89.77  \\
          & \checkmark     &       &       & 92.11  & 88.04  & 90.08  \\
    \checkmark     & \checkmark     &       &       & 92.44  & 87.90 & 90.17  \\
    \checkmark      & \checkmark      & \checkmark     &       & \textbf{92.56}  & 88.17  & 90.36  \\
    \checkmark      & \checkmark      &       & \checkmark     & 92.49  & 88.38  & 90.44  \\ \midrule
    \checkmark     & \checkmark     & \checkmark     & \checkmark     & 92.51 & \textbf{88.69} & \textbf{90.60} \\
  \bottomrule
\end{tabular}
\caption{Ablation study of components in the proposed method on BTCV 4-4 setting using average DSC (\%).}
\label{tab:ablation_component}
\end{table}

\subsection{Experimental Results}

\subsubsection{Results on BTCV dataset.}

Table~\ref{tab:main_result_BTCV} summarizes the performance comparison with SOTA CISS methods under different settings on the BTCV dataset. Our method consistently achieves the best overall DSC across all evaluated tasks. In particular, under the challenging 4-1 (5 steps) and 4-2 (3 steps) settings, our approach yields superior balance between old and new class performance, outperforming prior methods such as MiB and SDR. Notably, even in long incremental sequences, our method maintains strong segmentation accuracy without significant forgetting. Fig.~\ref{fig:visualization} presents a qualitative comparison between our approach and competing methods on the BTCV 4-4 task. The segmentation results further demonstrate the superiority of our method in both retaining old classes (\textit{e.g.}, left kidney) and learning new ones (\textit{e.g.}, liver).

\subsubsection{Results on WORD dataset.}

Table~\ref{tab:main_result_WORD} reports the performance of our method on various settings of the WORD dataset. Across all scenarios (4-4, 4-2, 2-2, 7-1), our approach consistently outperforms SOTA methods in average DSC. These results further demonstrate the effectiveness and generalization of our method on different datasets.

\subsection{Ablation Studies}

\subsubsection{Effectiveness of Each Component.}
To evaluate the contribution of each component in our framework, we conduct an ablation study on the BTCV 4-4 setting by incrementally adding individual loss terms: $\boldsymbol{\lambda}_{\text{ll}}$, $\boldsymbol{\lambda}_{\text{lg}}$, $\boldsymbol{\lambda}_{\text{orcd}}$, $\boldsymbol{\lambda}_{\text{crcd}}$. As shown in Table~\ref{tab:ablation_component}, adding each loss term individually improves performance over the baseline. Notably, combining $\boldsymbol{\lambda}_{\text{ll}}$ and $\boldsymbol{\lambda}_{\text{lg}}$ brings further gains, highlighting their complementarity. Both region-based distillation losses ($\boldsymbol{\lambda}_{\text{orcd}}$ and $\boldsymbol{\lambda}_{\text{crcd}}$) also contribute positively. The best overall DSC of 90.60 \% is achieved when all components are combined, confirming their effectiveness and synergy.

% Additionally, we provide ablation results regarding the weights of individual loss components in the supplementary materials.

\section{Conclusion}

In this paper, we mitigate the performance degradation of old knowledge caused by one-size-fits-all distillation and the neglect of old-class local representations in new data. Specifically, PGCD adaptively calibrates knowledge transfer across different regions and channels using class-specific prototype guidance, enhancing reliable old knowledge while suppressing misleading signals, thereby ensuring the learning of new categories. DAPD adopts an unbiased dual-path prototype alignment strategy to effectively preserve the performance of old classes. Extensive experiments on two public datasets and multiple incremental settings demonstrate the superiority of our approach over existing CISS methods.

\section{Acknowledgments}
This work was supported by the Major Program of the National Natural Science Foundation of China under Grant  62495064, National Natural Science Foundation of China under Grant 62476184, and Youth Innovation Research Team Project of Sichuan Province under Grant 2024NSFTD0051.

% \bigskip
% \noindent Thank you for reading these instructions carefully. We look forward to receiving your electronic files!

\bibliography{main}

\newpage
\appendix
\input{X_suppl}

\end{document}

%% file: X_suppl.tex
% \clearpage
% \setcounter{page}{1}
% \maketitlesupplementary

% \appendix
% \renewcommand{\thesection}{\arabic{section}}
% 这两句会让 标号 从 1 开始，和正文标号 隔离

\section{Notation}
Table~\ref{tab:notations} presents a comprehensive list of all symbols used in this paper along with their corresponding descriptions.

\section{Disscussion}
\subsection{Limitations and Future Work.}
Our method relies on pre-computed global prototypes, which can be sensitive to noisy annotations and domain shifts. In heterogeneous datasets or under cross-domain settings, such factors may lead to inaccurate prototype representations, potentially resulting in suboptimal calibration during distillation. In future work, we plan to develop noise-robust and domain-adaptive prototype estimation strategies, such as incorporating uncertainty weighting, prototype refinement via learned similarity metrics, or leveraging domain-invariant feature alignment. These improvements could further enhance the stability and generalization of our framework in challenging and diverse clinical scenarios.

% \looseness=-1

\section{More Details about Experimental Setup}

\subsection{Preprocessing.}

For preprocessing, we follow the standard pipeline established in prior work~\cite{zhang2023continual}. Specifically, for both BTCV and WORD, the Hounsfield Unit (HU) values are truncated to the range of [-175, 250] and normalized to the interval [0, 1]. All CT scans are resampled to a uniform voxel spacing of 1 $\times$ 1 $\times$ 3 $mm^3$. Subsequently, each dataset is randomly split into train and test sets using an 8:2 ratio, resulting in 24 train and 6 test scans for BTCV, and 96 train and 24 test scans for WORD.

\subsection{Implementation Details.}
For all experiments, we adopt Swin UNETR~\cite{hatamizadeh2021swin} as the backbone network, owing to its strong representation capability and effectiveness across various medical image segmentation tasks. The network is optimized using Stochastic Gradient Descent (SGD) with a momentum of 0.9. We use a fixed learning rate of 0.01 for both the initial and all subsequent incremental steps. For the BTCV dataset, we train the model with a batch size of 2 for 100 epochs, while for the WORD dataset, we use the same batch size of 2 and train for 50 epochs. During training, input volumes are randomly cropped to patches of size 96 $\times$ 96 $\times$ 96 from each CT scan. Following the prior work~\cite{cha2021ssul}, $\tau$ is set to 0.7.

\begin{table}[ht]
\centering
%\resizebox{.95\columnwidth}{!}{
\begin{tabular}{c | c | c | c | c | c}
    \toprule

    \multicolumn{6}{c}{$\boldsymbol{\lambda}_{\text{ll}}$} \\ \hline
    DSC (\%) & 0.1  & 0.5  & 1   & 5   & 10 \\ \hline
    Old   & 92.10  & 92.11  & 92.39  & 92.09  & 91.53  \\
    New   & 87.37  & 87.43  & 86.89  & 68.99  & 67.13  \\
    All   & 89.73  & \textbf{89.77} & 89.64  & 80.54  & 79.33  \\

    \midrule

    \multicolumn{6}{c}{$\boldsymbol{\lambda}_{\text{lg}}$} \\ \hline
    DSC (\%) & 0.1  & 0.5  & 1   & 5   & 10 \\ \hline
    Old   & 92.11  & 91.10  & 91.11  & 85.81  & 80.66  \\
    New   & 88.04  & 75.58  & 65.05  & 45.80  & 30.73  \\
    All   & \textbf{90.08} & 83.34  & 78.08  & 65.80  & 55.70  \\

    \midrule

    \multicolumn{6}{c}{$\boldsymbol{\lambda}_{\text{orcd}}$} \\ \hline
    DSC (\%) & 0.1  & 0.5  & 1   & 5   & 10 \\ \hline
    Old   & 92.16  & 91.87  & 92.29  & 92.19  & 91.51  \\
    New   & 88.29  & 87.46  & 88.38  & 87.65  & 86.79  \\
    All   & 90.22  & 89.66  & \textbf{90.33} & 89.92  & 89.15  \\

    \midrule
    
    \multicolumn{6}{c}{$\boldsymbol{\lambda}_{\text{crcd}}$} \\ \hline
    DSC (\%) & 0.1  & 0.5  & 1   & 5   & 10 \\ \hline
    Old   & 92.20  & 92.33  & 92.31  & 91.62  & 92.03  \\
    New   & 88.50  & 88.47  & 88.22  & 86.31  & 87.84  \\
    All   & 90.35  & \textbf{90.40} & 90.27  & 88.96  & 89.93  \\
    
    \bottomrule
\end{tabular}
\caption{Performance comparison across different weights ($\boldsymbol{\lambda}_{\text{ll}}$, $\boldsymbol{\lambda}_{\text{lg}}$, $\boldsymbol{\lambda}_{\text{orcd}}$, $\boldsymbol{\lambda}_{\text{crcd}}$) on BTCV 4-4 using average DSC (\%).}
\label{tab:ablation_weight}
\end{table}

\begin{figure}[ht]
\centering
\includegraphics[width=\columnwidth]{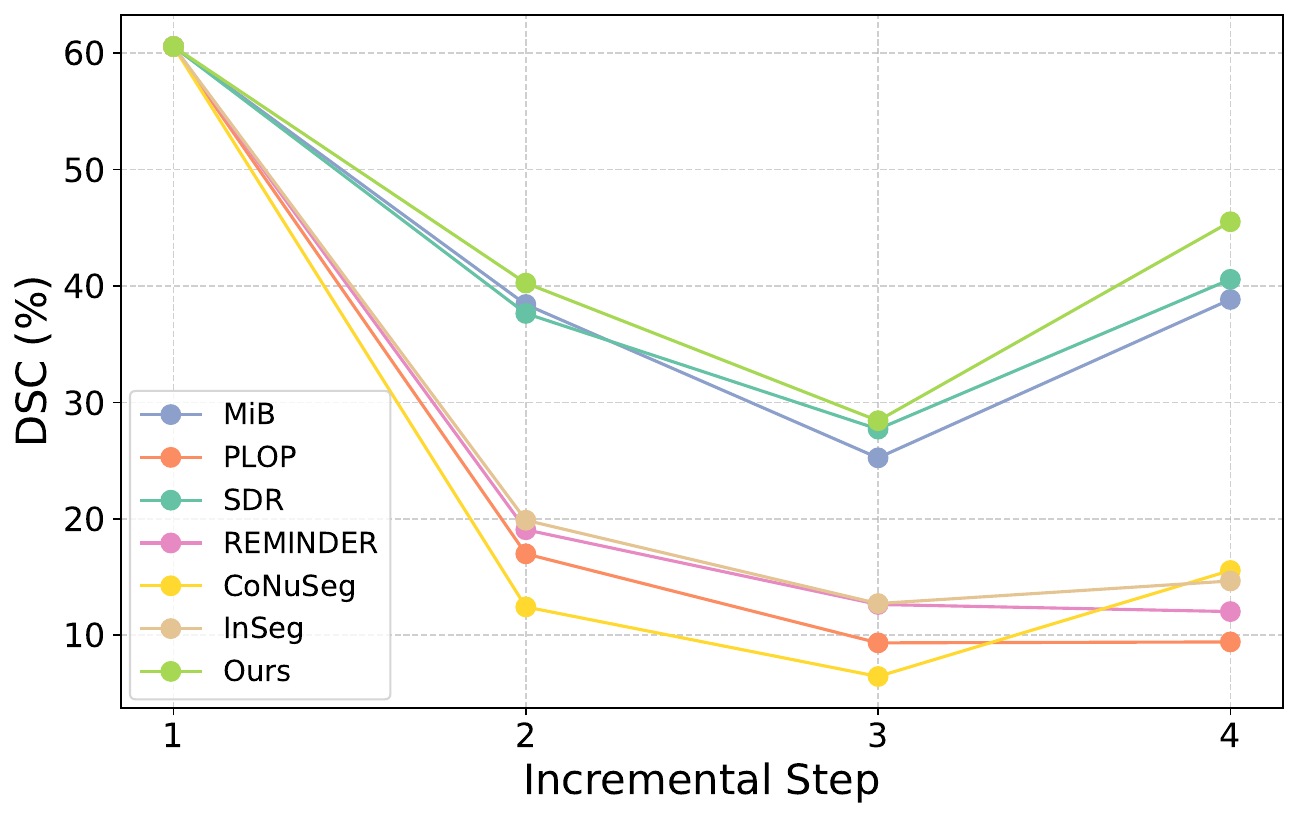} % Reduce the figure size so that it is slightly narrower than the column. Don't use precise values for figure width.This setup will avoid overfull boxes.
\caption{The average DSC (\%) of different methods at each incremental step on the BTCV 2-2 setting.}
\label{fig:broken_line}
\end{figure}

\begin{table*}[t]
\centering
%\resizebox{.95\columnwidth}{!}{
\begin{tabular}{lrrrrrrr}
    \toprule
    Method & \#Params (M) & \makecell[l]{Inference \\ Time (s)} & GFLOPs & \makecell[l]{Training \\ Time (s)} & \makecell[l]{Training CPU \\ Memory (GB)} & \makecell[l]{Training GPU \\ Memory (GiB)} & DSC (\%) \\
    \midrule
    LwF   & 62.19 & 2.45  & 329.84 & 34    & 1.75  & 14.72 & 87.11 \\
    ILT   & 62.19 & 2.45  & 329.84 & 35    & 1.78  & 14.72 & 31.26 \\
    MiB   & 62.19 & 2.45  & 329.84 & 35    & 1.79  & 14.72 & 89.12 \\
    PLOP  & 62.19 & 2.45  & 329.84 & 71    & 1.77  & 14.72 & 47.24 \\
    SDR   & 62.19 & 2.45  & 329.84 & 35    & 1.86  & 18.03 & 89.34 \\
    REMINDER & 62.19 & 2.45  & 329.84 & 71    & 1.8   & 14.72 & 28.37 \\
    CoNuSeg & 62.19 & 2.45  & 329.84 & 37    & 1.84  & 19.68 & 65.77 \\
    InSeg & 62.19 & 2.45  & 329.84 & 73    & 1.79  & 14.72 & 46.49 \\
    Ours  & 62.19 & 2.45  & 329.84 & 39    & 1.93  & 15.46 & 90.60 \\
    \bottomrule
\end{tabular}
\caption{Comparison of model complexity, computational cost, and segmentation performance across different CISS methods on BTCV 4-4 setting.}
\label{tab:efficiency}
\end{table*}

\begin{table*}[t]
\centering
\begin{tabular}{ccc}
\toprule
 & \textbf{Symbol} & \textbf{Description} \\ \hline

\multicolumn{1}{l}{\textbf{Space}} 
    & $\mathbb{R}^{H \times W \times D}$ & 3D volume space with height $H$, width $W$, depth $D$ \\
    & $\mathbb{R}^{V \times K}$ & Feature space with $V$ spatial locations and $K$ channels \\
    & $\mathbb{L}$ & Label space \\ \hline

\multicolumn{1}{l}{\textbf{Network}} 
    & $f_{\theta}^{t}$ & Segmentation model at step $t$ with parameters $\theta$ \\
    & $f_{\theta}^{t-1}$ & Segmentation model from previous step \\
     \hline

\multicolumn{1}{l}{\textbf{Sets}} 
    & $\mathcal{D}^t$ & Training dataset at step $t$ \\
    & $\mathcal{D}^{1:t-1}$ & Datasets from previous steps \\
    & $\mathcal{C}^t$ & Class set introduced at step $t$ \\
    & $\mathcal{C}^{1:t}$ & All classes seen from step 1 to $t$ \\
    & $\mathcal{C}^{1:t-1}$ & Old classes from previous steps \\
    & $\mathcal{C}^{t+1:T}$ & Future classes not yet introduced \\ \hline

\multicolumn{1}{l}{\textbf{Samples}} 
    & $\mathbf{X}_n$ & $n$-th 3D medical image volume \\
    & $\mathbf{Y}_n$ & Ground truth label map of $\mathbf{X}_n$ \\
    & $\mathbf{\hat{Y}}$ & Model prediction (logits or probabilities) \\
    & $\mathbf{\tilde{Y}}$ & Pseudo label \\
    & $\mathbf{F}^t, \mathbf{F}^{t-1}$ & Feature maps from current and old models \\
    & $\hat{\mathbf{S}}^t, \hat{\mathbf{S}}^{t-1}$ & Softmax outputs of current and old models \\
    & $\mathbf{M}_i^{t}, \mathbf{M}_i^{t-1}$ & Current and old region mask \\
    & $\mathbf{a}^{t-1}, \mathbf{a}^{t}$ & Current and old region affinity \\
    & $N_t$ & Number of samples at step $t$ \\
    & $V$ & Number of spatial locations (voxels) \\
    & $K$ & Feature channel dimension \\
    & $B$ & Batch size \\
    & $H, W, D$ & Height, width, depth \\ \hline

\multicolumn{1}{l}{\textbf{Prototypes}} 
    & $\mathbf{\hat{p}}_c$ & Local prototype of class $c$ \\
    & $\mathbf{p}_c$ & Global prototype of class $c$ (moving average) \\
    & $\mathbf{\hat{q}}_c$ & Unbiased transformed prototype of class $c$ \\
    & $\mathbf{F}_c$ & Feature vectors belonging to class $c$ \\
    & $N_c^{\text{pre}}$ & Number of observed voxels of class $c$ before current batch \\ 
    & $\mathbf{p}_c^{\text{pre}}$ & Cumulative global prototypes of class $c$ before current batch \\ 
    \hline

\multicolumn{1}{l}{\textbf{Miscellaneous}} 
    & $\mathcal{L}_{\text{ce}}$ & Unbiased cross-entropy loss for current classes \\
    & $\mathcal{L}_{\text{orcd}}$ & Old Region Calibrated Distillation loss \\
    & $\mathcal{L}_{\text{crcd}}$ & Current Region Calibrated Distillation loss \\
    & $\mathcal{L}_{\text{dapd}}$ & Dual-Aligned Prototype Distillation loss \\
    & $\mathcal{L}_{\text{pd}}$ & Prototype distillation loss \\
    & $\boldsymbol{\lambda}_{\ast}$ & Weight of the corresponding loss term \\
    & $\mathbb{I}[\cdot]$ & Indicator function \\
    & $\|\cdot\|_2$ & L2 norm (Euclidean norm) \\
    & KL$(\cdot\parallel\cdot)$ & Kullback-Leibler divergence \\
    & $u_i$ & Entropy-based uncertainty of voxel $i$ \\
    & $\tau$ & Uncertainty threshold \\
\bottomrule
\end{tabular}
\caption{Summary of notations used in this paper.}
\label{tab:notations}
\end{table*}

\section{More Experimental Results}

\subsection{Ablation of Hyperparameters.}

To investigate the impact of different loss components, we conduct an ablation study on the BTCV 4-4 setting by varying the weights of four loss terms: $\boldsymbol{\lambda}_{\text{ll}}$, $\boldsymbol{\lambda}_{\text{lg}}$, $\boldsymbol{\lambda}_{\text{orcd}}$, $\boldsymbol{\lambda}_{\text{crcd}}$. Each weight is varied from 0.1 to 10, and we report the average DSC (\%) on ``Old'', ``New'', and ``All'' classes. As shown in Table~\ref{tab:ablation_weight}, the model achieves the highest DSC score when the weights $\boldsymbol{\lambda}_{\text{ll}}$ and $\boldsymbol{\lambda}_{\text{lg}}$ are set to 0.5 and 0.1, respectively. When the values of $\boldsymbol{\lambda}_{\text{ll}}$ exceed 5 and $\boldsymbol{\lambda}_{\text{lg}}$ exceed 0.5, the performance on new classes significantly degrades and overall accuracy drops, indicating that overly strong alignment with old prototypes may hinder the learning of new knowledge. 

In contrast, $\boldsymbol{\lambda}_{\text{orcd}}$ and $\boldsymbol{\lambda}_{\text{crcd}}$ exhibit more stable behaviors across a wide range of values, showing robustness and consistency in maintaining a favorable trade-off between old and new class performance. These results demonstrate the necessity of balancing loss weights to effectively stabilize the model in continual learning scenarios.

\subsection{More Results on BTCV Dataset.}

Fig.~\ref{fig:broken_line} presents the DSC (\%) evolution over incremental steps for different continual segmentation methods. While all approaches maintain stable scores at the initial step, most competing methods experience a substantial performance drop as more incremental steps are introduced. For example, PLOP, SDR, and REMINDER exhibit over 20 \% DSC degradation from the first to the last step, highlighting their limited ability to retain previously learned knowledge under long incremental sequences. In contrast, our method sustains consistently higher DSC across all steps, with only a marginal decrease, indicating superior stability and robustness against catastrophic forgetting. These results confirm that our method achieves a better balance between stability and plasticity throughout the incremental learning process.

\subsection{Comparison of Model Efficiency.}

To comprehensively evaluate the efficiency and performance of different CISS methods, we conduct a fair comparison under the same experimental setup. Table~\ref{tab:efficiency} summarizes the results in terms of model parameters, inference and training time, memory consumption, and segmentation accuracy (DSC). As shown, our method achieves the highest DSC (90.60 \%) while maintaining comparable computational cost, demonstrating an effective trade-off between accuracy and efficiency.

%% file: Formatting-Instructions-LaTeX-2026.bbl
\begin{thebibliography}{46}
\providecommand{\natexlab}[1]{#1}

\bibitem[{Aljundi et~al.(2019)Aljundi, Lin, Goujaud, and Bengio}]{aljundi2019gradient}
Aljundi, R.; Lin, M.; Goujaud, B.; and Bengio, Y. 2019.
\newblock Gradient based sample selection for online continual learning.
\newblock \emph{Advances in neural information processing systems}.

\bibitem[{Bang et~al.(2021)Bang, Kim, Yoo, Ha, and Choi}]{bang2021rainbow}
Bang, J.; Kim, H.; Yoo, Y.; Ha, J.-W.; and Choi, J. 2021.
\newblock Rainbow memory: Continual learning with a memory of diverse samples.
\newblock In \emph{CVPR}.

\bibitem[{Bauer et~al.(2013)Bauer, Wiest, Nolte, and Reyes}]{bauer2013survey}
Bauer, S.; Wiest, R.; Nolte, L.-P.; and Reyes, M. 2013.
\newblock A survey of MRI-based medical image analysis for brain tumor studies.
\newblock \emph{Physics in Medicine \& Biology}.

\bibitem[{Burgos et~al.(2017)Burgos, Guerreiro, McClelland, Presles, Modat, Nill, Dearnaley, Desouza, Oelfke, Knopf et~al.}]{burgos2017iterative}
Burgos, N.; Guerreiro, F.; McClelland, J.; Presles, B.; Modat, M.; Nill, S.; Dearnaley, D.; Desouza, N.; Oelfke, U.; Knopf, A.-C.; et~al. 2017.
\newblock Iterative framework for the joint segmentation and CT synthesis of MR images: application to MRI-only radiotherapy treatment planning.
\newblock \emph{Physics in Medicine \& Biology}.

\bibitem[{Cermelli et~al.(2020)Cermelli, Mancini, Bulo, Ricci, and Caputo}]{cermelli2020modeling}
Cermelli, F.; Mancini, M.; Bulo, S.~R.; Ricci, E.; and Caputo, B. 2020.
\newblock Modeling the background for incremental learning in semantic segmentation.
\newblock In \emph{CVPR}.

\bibitem[{Cha et~al.(2021)Cha, Yoo, Moon et~al.}]{cha2021ssul}
Cha, S.; Yoo, Y.; Moon, T.; et~al. 2021.
\newblock Ssul: Semantic segmentation with unknown label for exemplar-based class-incremental learning.
\newblock \emph{Advances in neural information processing systems}.

\bibitem[{Cong et~al.(2024)Cong, Cong, Liu, and Sun}]{cong2024cs}
Cong, W.; Cong, Y.; Liu, Y.; and Sun, G. 2024.
\newblock Cs 2 K: Class-Specific and Class-Shared Knowledge Guidance for Incremental Semantic Segmentation.
\newblock In \emph{ECCV}.

\bibitem[{Douillard et~al.(2021)Douillard, Chen, Dapogny, and Cord}]{douillard2021plop}
Douillard, A.; Chen, Y.; Dapogny, A.; and Cord, M. 2021.
\newblock Plop: Learning without forgetting for continual semantic segmentation.
\newblock In \emph{CVPR}.

\bibitem[{Douillard et~al.(2020)Douillard, Cord, Ollion, Robert, and Valle}]{douillard2020podnet}
Douillard, A.; Cord, M.; Ollion, C.; Robert, T.; and Valle, E. 2020.
\newblock Podnet: Pooled outputs distillation for small-tasks incremental learning.
\newblock In \emph{ECCV}.

\bibitem[{French(1999)}]{french1999catastrophic}
French, R.~M. 1999.
\newblock Catastrophic forgetting in connectionist networks.
\newblock \emph{Trends in cognitive sciences}.

\bibitem[{Hatamizadeh et~al.(2021)Hatamizadeh, Nath, Tang, Yang, Roth, and Xu}]{hatamizadeh2021swin}
Hatamizadeh, A.; Nath, V.; Tang, Y.; Yang, D.; Roth, H.~R.; and Xu, D. 2021.
\newblock Swin unetr: Swin transformers for semantic segmentation of brain tumors in mri images.
\newblock In \emph{International MICCAI brainlesion workshop}.

\bibitem[{Iscen et~al.(2020)Iscen, Zhang, Lazebnik, and Schmid}]{iscen2020memory}
Iscen, A.; Zhang, J.; Lazebnik, S.; and Schmid, C. 2020.
\newblock Memory-efficient incremental learning through feature adaptation.
\newblock In \emph{ECCV}.

\bibitem[{Kirkpatrick et~al.(2017)Kirkpatrick, Pascanu, Rabinowitz, Veness, Desjardins, Rusu, Milan, Quan, Ramalho, Grabska-Barwinska et~al.}]{kirkpatrick2017overcoming}
Kirkpatrick, J.; Pascanu, R.; Rabinowitz, N.; Veness, J.; Desjardins, G.; Rusu, A.~A.; Milan, K.; Quan, J.; Ramalho, T.; Grabska-Barwinska, A.; et~al. 2017.
\newblock Overcoming catastrophic forgetting in neural networks.
\newblock \emph{Proceedings of the national academy of sciences}.

\bibitem[{Landman et~al.(2015)Landman, Xu, Igelsias, Styner, Langerak, and Klein}]{landman2015miccai}
Landman, B.; Xu, Z.; Igelsias, J.; Styner, M.; Langerak, T.; and Klein, A. 2015.
\newblock Miccai multi-atlas labeling beyond the cranial vault--workshop and challenge.
\newblock In \emph{Proc. MICCAI Multi-Atlas Labeling Beyond Cranial Vault—Workshop Challenge}.

\bibitem[{Li and Hoiem(2017)}]{li2017learning}
Li, Z.; and Hoiem, D. 2017.
\newblock Learning without forgetting.
\newblock \emph{IEEE transactions on pattern analysis and machine intelligence}.

\bibitem[{Lin, Wang, and Zhang(2022)}]{lin2022continual}
Lin, Z.; Wang, Z.; and Zhang, Y. 2022.
\newblock Continual semantic segmentation via structure preserving and projected feature alignment.
\newblock In \emph{ECCV}.

\bibitem[{Luo et~al.(2022)Luo, Liao, Xiao, Chen, Song, Zhang, Li, Metaxas, Wang, and Zhang}]{luo2022word}
Luo, X.; Liao, W.; Xiao, J.; Chen, J.; Song, T.; Zhang, X.; Li, K.; Metaxas, D.~N.; Wang, G.; and Zhang, S. 2022.
\newblock WORD: A large scale dataset, benchmark and clinical applicable study for abdominal organ segmentation from CT image.
\newblock \emph{Medical Image Analysis}.

\bibitem[{Masana et~al.(2022)Masana, Liu, Twardowski, Menta, Bagdanov, and Van De~Weijer}]{masana2022class}
Masana, M.; Liu, X.; Twardowski, B.; Menta, M.; Bagdanov, A.~D.; and Van De~Weijer, J. 2022.
\newblock Class-incremental learning: survey and performance evaluation on image classification.
\newblock \emph{IEEE Transactions on Pattern Analysis and Machine Intelligence}.

\bibitem[{McCloskey and Cohen(1989)}]{mccloskey1989catastrophic}
McCloskey, M.; and Cohen, N.~J. 1989.
\newblock Catastrophic interference in connectionist networks: The sequential learning problem.
\newblock In \emph{Psychology of learning and motivation}.

\bibitem[{Michieli and Zanuttigh(2019)}]{michieli2019incremental}
Michieli, U.; and Zanuttigh, P. 2019.
\newblock Incremental learning techniques for semantic segmentation.
\newblock In \emph{ICCVW}.

\bibitem[{Michieli and Zanuttigh(2021)}]{michieli2021continual}
Michieli, U.; and Zanuttigh, P. 2021.
\newblock Continual semantic segmentation via repulsion-attraction of sparse and disentangled latent representations.
\newblock In \emph{CVPR}.

\bibitem[{Phan et~al.(2022)Phan, Phung, Tran-Thanh, Bouzerdoum et~al.}]{phan2022class}
Phan, M.~H.; Phung, S.~L.; Tran-Thanh, L.; Bouzerdoum, A.; et~al. 2022.
\newblock Class similarity weighted knowledge distillation for continual semantic segmentation.
\newblock In \emph{CVPR}.

\bibitem[{Qi, Wu, and Chan(2023)}]{qi2023mdf}
Qi, W.; Wu, H.-C.; and Chan, S.-C. 2023.
\newblock Mdf-net: A multi-scale dynamic fusion network for breast tumor segmentation of ultrasound images.
\newblock \emph{IEEE Transactions on Image Processing}, 32.

\bibitem[{Qi, Wu, and Chan(2024)}]{qi2024gradient}
Qi, W.; Wu, J.; and Chan, S. 2024.
\newblock Gradient-aware for class-imbalanced semi-supervised medical image segmentation.
\newblock In \emph{European Conference on Computer Vision}.

\bibitem[{Robins(1995)}]{robins1995catastrophic}
Robins, A. 1995.
\newblock Catastrophic forgetting, rehearsal and pseudorehearsal.
\newblock \emph{Connection Science}.

\bibitem[{Shin et~al.(2017)Shin, Lee, Kim, and Kim}]{shin2017continual}
Shin, H.; Lee, J.~K.; Kim, J.; and Kim, J. 2017.
\newblock Continual learning with deep generative replay.
\newblock \emph{Advances in neural information processing systems}.

\bibitem[{Snell, Swersky, and Zemel(2017)}]{snell2017prototypical}
Snell, J.; Swersky, K.; and Zemel, R. 2017.
\newblock Prototypical networks for few-shot learning.
\newblock \emph{Advances in neural information processing systems}.

\bibitem[{Tang et~al.(2024)Tang, Lu, Xu, Wu, Hu, Zhang, Cheng, Ge, Chen, and Tsung}]{tang2024incremental}
Tang, J.; Lu, H.; Xu, X.; Wu, R.; Hu, S.; Zhang, T.; Cheng, T.~W.; Ge, M.; Chen, Y.-C.; and Tsung, F. 2024.
\newblock An Incremental Unified Framework for Small Defect Inspection.
\newblock In \emph{ECCV}.

\bibitem[{Wang et~al.(2022)Wang, Zhou, Ye, and Zhan}]{wang2022foster}
Wang, F.-Y.; Zhou, D.-W.; Ye, H.-J.; and Zhan, D.-C. 2022.
\newblock Foster: Feature boosting and compression for class-incremental learning.
\newblock In \emph{ECCV}.

\bibitem[{Wang, Wu, and Qin(2024)}]{wang2024incremental}
Wang, H.; Wu, H.; and Qin, J. 2024.
\newblock Incremental Nuclei Segmentation from Histopathological Images via Future-class Awareness and Compatibility-inspired Distillation.
\newblock In \emph{CVPR}.

\bibitem[{Wieser et~al.(2017)Wieser, Cisternas, Wahl, Ulrich, Stadler, Mescher, M{\"u}ller, Klinge, Gabrys, Burigo et~al.}]{wieser2017development}
Wieser, H.-P.; Cisternas, E.; Wahl, N.; Ulrich, S.; Stadler, A.; Mescher, H.; M{\"u}ller, L.-R.; Klinge, T.; Gabrys, H.; Burigo, L.; et~al. 2017.
\newblock Development of the open-source dose calculation and optimization toolkit matRad.
\newblock \emph{Medical physics}.

\bibitem[{Wu et~al.(2023)Wu, Wang, Zhao, Chen, and Qin}]{wu2023continual}
Wu, H.; Wang, Z.; Zhao, Z.; Chen, C.; and Qin, J. 2023.
\newblock Continual nuclei segmentation via prototype-wise relation distillation and contrastive learning.
\newblock \emph{IEEE Transactions on Medical Imaging}.

\bibitem[{Wu, Xu, and Tong(2024)}]{wu2024continual}
Wu, X.; Xu, Z.; and Tong, R. K.-y. 2024.
\newblock Continual learning in medical image analysis: A survey.
\newblock \emph{Computers in Biology and Medicine}.

\bibitem[{Xiaogang~Xu and Jia(2021)}]{xu2021ddcat}
Xiaogang~Xu, H.~Z.; and Jia, J. 2021.
\newblock Dynamic Divide-and-Conquer Adversarial Training for Robust Semantic Segmentation.
\newblock In \emph{ICCV}.

\bibitem[{Xu et~al.(2022{\natexlab{a}})Xu, Zhao, Torr, and Jia}]{xu2022general}
Xu, X.; Zhao, H.; Torr, P.; and Jia, J. 2022{\natexlab{a}}.
\newblock General adversarial defense against black-box attacks via pixel level and feature level distribution alignments.
\newblock \emph{arXiv preprint arXiv:2212.05387}.

\bibitem[{Xu et~al.(2022{\natexlab{b}})Xu, Zhao, Vineet, Lim, and Torralba}]{xu2022mtformer}
Xu, X.; Zhao, H.; Vineet, V.; Lim, S.-N.; and Torralba, A. 2022{\natexlab{b}}.
\newblock MTFormer: Multi-Task Learning via Transformer and Cross-Task Reasoning.
\newblock In \emph{ECCV}.

\bibitem[{Yan, Xie, and He(2021)}]{yan2021dynamically}
Yan, S.; Xie, J.; and He, X. 2021.
\newblock Der: Dynamically expandable representation for class incremental learning.
\newblock In \emph{CVPR}.

\bibitem[{Zhang and Gao(2024)}]{zhang2024background}
Zhang, A.; and Gao, G. 2024.
\newblock Background adaptation with residual modeling for exemplar-free class-incremental semantic segmentation.
\newblock In \emph{European Conference on Computer Vision}.

\bibitem[{Zhang et~al.(2023)Zhang, Li, Chen, Yuille, Liu, and Zhou}]{zhang2023continual}
Zhang, Y.; Li, X.; Chen, H.; Yuille, A.~L.; Liu, Y.; and Zhou, Z. 2023.
\newblock Continual learning for abdominal multi-organ and tumor segmentation.
\newblock In \emph{International conference on medical image computing and computer-assisted intervention}.

\bibitem[{Zhang et~al.(2017)Zhang, Xie, Xing, McGough, and Yang}]{zhang2017mdnet}
Zhang, Z.; Xie, Y.; Xing, F.; McGough, M.; and Yang, L. 2017.
\newblock Mdnet: A semantically and visually interpretable medical image diagnosis network.
\newblock In \emph{CVPR}.

\bibitem[{Zhou et~al.(2024)Zhou, Wang, Qi, Ye, Zhan, and Liu}]{zhou2024class}
Zhou, D.-W.; Wang, Q.-W.; Qi, Z.-H.; Ye, H.-J.; Zhan, D.-C.; and Liu, Z. 2024.
\newblock Class-incremental learning: A survey.
\newblock \emph{IEEE Transactions on Pattern Analysis and Machine Intelligence}.

\bibitem[{Zhou et~al.(2022)Zhou, Wang, Ye, and Zhan}]{zhou2022model}
Zhou, D.-W.; Wang, Q.-W.; Ye, H.-J.; and Zhan, D.-C. 2022.
\newblock A model or 603 exemplars: Towards memory-efficient class-incremental learning.
\newblock \emph{arXiv preprint arXiv:2205.13218}.

\bibitem[{Zhu et~al.(2025{\natexlab{a}})Zhu, Wu, Gao, Wang, Yang, and Sang}]{zhu2025adaptive}
Zhu, G.; Wu, D.; Gao, C.; Wang, R.; Yang, W.; and Sang, N. 2025{\natexlab{a}}.
\newblock Adaptive prototype replay for class incremental semantic segmentation.
\newblock In \emph{AAAI}.

\bibitem[{Zhu et~al.(2025{\natexlab{b}})Zhu, Wu, Xu, Yu, Song, Yi, Li, and Hu}]{zhu2025exploiting}
Zhu, S.; Wu, J.; Xu, X.; Yu, C.; Song, Y.; Yi, Z.; Li, G.; and Hu, J. 2025{\natexlab{b}}.
\newblock Exploiting Unlabeled Structures through Task Consistency Training for Versatile Medical Image Segmentation.
\newblock \emph{arXiv preprint arXiv:2509.04732}.

\bibitem[{Zhu et~al.(2025{\natexlab{c}})Zhu, Yu, Qi, Wu, Song, Li, Yi, Xu, and Hu}]{zhu2025prime}
Zhu, S.; Yu, C.; Qi, W.; Wu, J.; Song, Y.; Li, G.; Yi, Z.; Xu, X.; and Hu, J. 2025{\natexlab{c}}.
\newblock PRIME: Prototype-Driven Class Incremental Learning for Medical Image Segmentation.
\newblock In \emph{Proceedings of the 33rd ACM International Conference on Multimedia}.

\bibitem[{Zhu et~al.(2025{\natexlab{d}})Zhu, Yu, Yi, and Hu}]{zhu2025visual}
Zhu, S.; Yu, C.; Yi, Z.; and Hu, J. 2025{\natexlab{d}}.
\newblock Visual prompt-driven universal model for medical image segmentation in radiotherapy.
\newblock \emph{Knowledge-Based Systems}.

\end{thebibliography}
